\pdfoutput=1

\documentclass[11pt]{article}

\usepackage{EMNLP2022}

\usepackage{times}
\usepackage{latexsym}
\usepackage{graphicx}
\usepackage{amsmath}
\usepackage{algorithm}
\usepackage{algpseudocode}
\usepackage{mathrsfs}
\usepackage{multirow}
\usepackage{booktabs}
\usepackage{enumitem}
\usepackage{bbm}

\usepackage[T1]{fontenc}

\usepackage[utf8]{inputenc}

\usepackage{microtype}

\usepackage{inconsolata}
\usepackage{fontawesome5}

\usepackage{listings}
\newcommand{\nop}[1]{}

\title{Limitations of Language Models in Arithmetic and Symbolic Induction\thanks{\text{     } The first two authors (Jing and Hong) contributed equally to this work.}}

\author{Jing Qian$^*$, Hong Wang$^*$, Zekun Li, Shiyang Li, Xifeng Yan \\
        University of California, Santa Barbara\\
        \tt \{jing\_qian, hongwang600, zekunli, shiyangli, xyan\}@cs.ucsb.edu
}
\begin{document}
\maketitle
\begin{abstract}
Recent work has shown that large pretrained Language Models (LMs) can not only perform remarkably well on a range of Natural Language Processing (NLP) tasks but also start improving on reasoning tasks such as arithmetic induction, symbolic manipulation, and commonsense reasoning with increasing size of models~\cite{cot,palm}. However, it is still unclear what the underlying capabilities of these LMs are. Surprisingly, we find that these models have limitations on certain basic symbolic manipulation tasks such as copy, reverse, and addition. When the total number of symbols or repeating symbols increases, the model performance drops quickly. We investigate the potential causes behind this phenomenon and examine a set of possible methods, including explicit positional markers, fine-grained computation steps, and LMs with callable programs.  Experimental results show that none of these techniques can solve the simplest addition induction problem completely.  In the end, we introduce LMs with tutor, which demonstrates every single step of teaching.  LMs with tutor is able to deliver 100\% accuracy in situations of OOD and repeating symbols, shedding new insights on the boundary of large LMs in induction.
\end{abstract}

\section{Introduction}
Transformer-based large pretrained Language Models, such as GPT3 and T5~\cite{transformer, gpt3, t5}, have been widely used as few-shot learners in many NLP tasks. Recent work even finds these models can achieve state-of-the-art performance in arithmetic  and symbolic reasoning~\cite{ scratchpad,cot}. 
Although these models exhibit surprisingly impressive capabilities in complex arithmetic reasoning tasks, such as MultiArith~\cite{multiarith} and GSM8k~\cite{gsm8k}, it has also been pointed out that they tend to make certain calculation errors and perform significantly worse when the number of math operations increases in equations~\cite{cot}. \citet{gpt3} find that GPT3 displays strong proficiency in 2-digit arithmetic addition, but struggles in arithmetic addition on numbers with more than three digits. \citet{investigating2021} also observe that the fine-tuned T5 model can not correctly add or subtract arbitrarily long numbers.  Larger models might perform better on the testing data, but worse on numbers that are longer than the training data (out-of-distribution, OOD) \cite{investigating2021}. However, even with the largest T5 model they experimented, the out-of-distribution (OOD) accuracy is not as high as the in-distribution accuracy, and increasing the training data does not improve OOD generalization beyond a critical amount.

\begin{figure}[!t]
\centering
\includegraphics[width=0.4\textwidth]{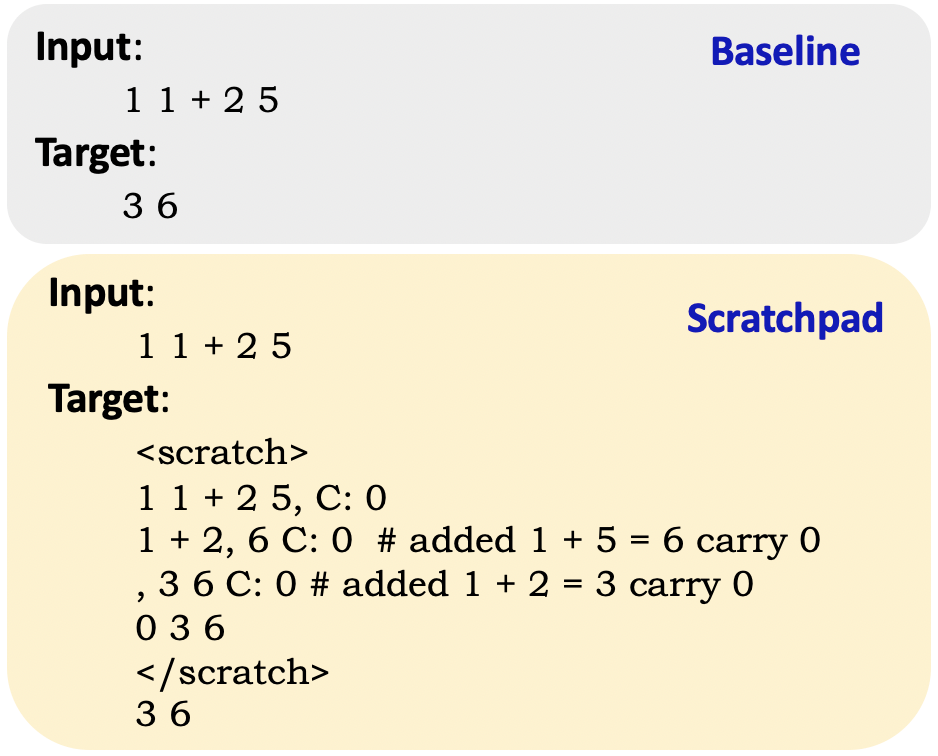}
 \caption{Examples of addition: the baseline setting (top) and Scratchpad \cite{scratchpad} with intermediate computation steps (bottom).  A similar method with more detailed demonstration is introduced in \cite{Recchia-2021}. }
\label{fig:intro-addition-example}
\end{figure}
\begin{figure*}[!t]
\centering
\includegraphics[width=0.98\textwidth]{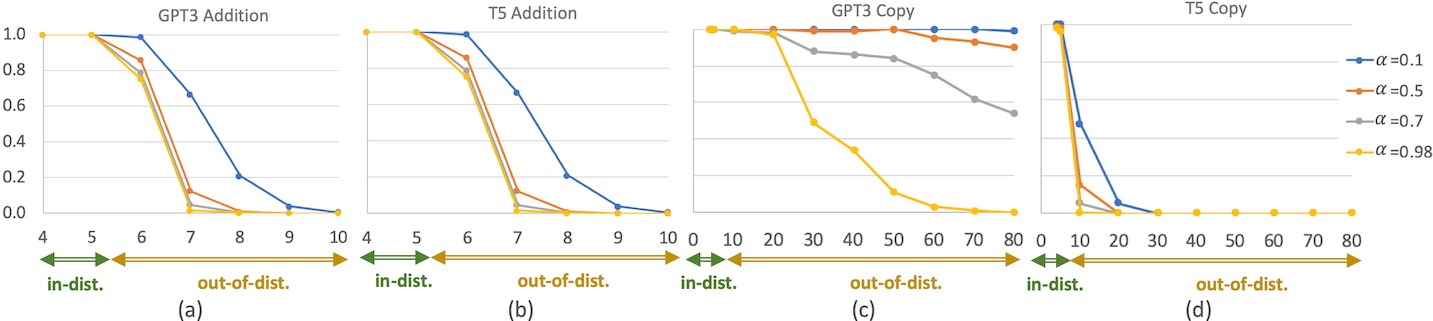}
\caption{The horizontal axis is the number of digits and the vertical axis is the accuracy. The prompts for GPT3 consist of 4 examples. The T5 models are trained on 1-5 digits of up to 2,000 examples and each training example consists of random numbers in the format of \texttt{2 4 1}. In-dist: in-distribution. Out-of-dist.: out-of-distribution (OOD).  In-distribution
refers to training on up to k-digit numbers and
testing on up to k-digit numbers while out-of-
distribution refers to training on up to k-digit numbers and testing on numbers with more digits.   $\alpha$ indicates the repetition level of the examples. An example $x_1\cdots x_n$ with $n$ digits are sampled with the next digit probability $p(x_{i+1}|x_{i})= \alpha$, when  $x_{i+1}=x_{i}$; otherwise,  $(1-\alpha)/9$. Larger $\alpha$ indicates a higher repetition level. 
}
\label{fig:intro-repeat}
\end{figure*}

Figure~\ref{fig:intro-addition-example} shows two possible addition exemplars for LMs. Addition can be considered as a basic arithmetic operation and a simple symbolic manipulation task.  The scratchpad version gives more details on how humans do basic arithmetic.  \citet{scratchpad} show that with more fine-grained demonstrations, the accuracy of addition can be improved dramatically with fine-tuning. Yet, it still can not achieve 100\% on OOD data, even with thousands of training data points provided. Figure \ref{fig:intro-repeat} shows the performance of GPT-3 and T5 on addition using the scratchpad version of training data. The problem becomes more severe when there are repeating digits in the addition operands.  

As the performance drops with repeating digits, we suspect that LMs might not handle the repeating symbols well. Figure \ref{fig:intro-repeat} illustrates the performance of GPT-3 and T5 on the copy task, one of the simplest symbolic manipulation operations. GPT-3 and T5 still can not perform well on OOD. We further do a preliminary experiment where a T5 model is fine-tuned using the data containing repeating numbers of up to 80 digits, T5 still can not achieve 100\% in-distribution accuracy on long repeating digits. The results indicate that there are two problems intervening: Transformers are not good at handling repeating symbols and OOD generalization. The repeating symbols can also be a problem even for in-distribution data. 

Why do large pretrained LMs that can do complex language generation fail on a simple symbolic manipulation task?  

In this paper, we investigate the potential causes behind this phenomenon. We believe that overcoming the aforementioned limitations is of critical importance for the future application of Transformer-based LMs to reasoning-intensive tasks.  What are the necessary steps to take to significantly improve or even approach 100\% accuracy on these simple but fundamentally important induction tasks?  We examine a set of possible mitigation solutions including fine-grained computation steps, positional markers, and LMs with callable programs. 

Since incorporating computation steps improves the OOD generalization in arithmetic addition~\cite{scratchpad}, one possible direction is to provide more fine-grained computation steps in the fine-tuning data or the few-shot prompt. However, it may not be sufficient to alleviate the problem of repeating numbers. When a human does addition, the position of each digit is used to differentiate the repeating digits. However, the self-attention mechanism in the Transformer may not tell which ``1'' is referred to in the input. This prompts us to explore using positional markers to differentiate the important tokens. Using these two methods to augment the reasoning process, we find that the performance of pretrained LMs still can not reach satisfying results. Then we resort to a method where the copy operation is implemented as a primitive function and explore whether the LM can further boost its performance.  

We experiment with three symbolic manipulation tasks: copying, reversing, and addition. Experimental results show that although generalization in these symbolic manipulation tasks is straightforward for humans, it is still challenging for LMs, and none of these mitigation methods fully solves the problems. In the end, we introduce LMs with tutor which demonstrates every single step of teaching, pinpointing where these digits come from.  LMs with tutor is able to deliver 100\% accuracy in situations of OOD and repeated symbols. In this design, LMs are used to generate actions that mimic operations in multiple tape Turing machines, rather than the intermediate results. These actions generate the intermediate results on tapes.   We hope this could shed light on the capability of Transformer-based LMs in addition to providing large training datasets or scaling up the size of these models. 

To conclude, our main contributions are:
\begin{itemize}
    
    \item We identify a set of simple symbolic manipulation tasks and uncover the limitations of the LMs in arithmetic and symbolic induction.
    
    \item We examine a set of potential techniques including positional markers, fine-grained computation steps, and LMs with callable programs.  Though these techniques could mitigate the limitations of the LMs, none of them can completely solve the generalization problem.
    
    \item Finally, we demonstrate that LMs with tutor is able to deliver 100\% accuracy in situations of OOD and repeated symbols. Our analysis could inspire new thoughts to overcome the limitation of LMs in symbolic manipulation.
    
\end{itemize}

\section{Related Work}
Previous research in using LMs for symbolic induction improves the model performance in the following three directions.

\textbf{Large Pretrained Language Models}:  \citet{gpt3} show that GPT3 exhibits  strong proficiency on 2-digit addition and subtraction using simply few-shot prompting, without any task-specific training. Furthermore, the larger the LM, the better the performance. Following GPT3, \citet{palm} further scale the Transformer-based LMs to a 540-billion parameter model, called Pathways Language Model (PaLM). Same as \citet{gpt3}, \citet{palm} find that scaling the LMs consistently results in better  arithmetic reasoning ability with few-shot prompting. However, the reasoning ability of the large LMs is still limited. GPT3 struggles with 3-digit arithmetic and with direct prompting, even 540B PaLM can not achieve high performance on complex tasks requiring multi-step reasoning. Therefore \citet{cot} propose to augment the large pretrained LMs with the following prompting method.

\textbf{Chain-of-Thought Prompting}: This prompting method provides a few chain-of-thought demonstrations, which is a series of intermediate reasoning steps, as exemplars in the prompting. Therefore, given a complex reasoning task, the model is allowed to calculate the intermediate results step-by-step before generating the final answer. With chain-of-thought prompting, a complex reasoning task is decomposed into a list of simple operations and LMs can 
derive these operations one by one. \citet{kim-etal-2022-ept} adopt faithful explanations that accurately represent the reasoning process behind solving a  math word problem.   \citet{cot} show that combining chain-of-thought prompting and a sufficiently large LM, 540B PaLM, can significantly improve the LMs' reasoning ability on complex tasks, such as math word problems.

\textbf{Fine-tuning with Large Training Datasets}:
Instead of few-shot prompting, another direction is to fine-tune large LMs with a sufficient amount of training data. \citet{investigating2021} fine-tune T5 with different ways of representing numbers, but even with the best-performing representation, the fine-tuned model can not achieve as good accuracy on out-of-distribution testing examples as in-distribution testing examples. \nop{Here in-distribution refers to training on up to k-digit numbers and testing on up to k-digit numbers while out-of-distribution refers to training on up to k-digit numbers and testing on numbers with more digits.} \citet{scratchpad} propose to use Scratchpad to improve the out-of-distribution accuracy. Scratchpad combines step-by-step reasoning with fine-tuning. The training examples include the intermediate steps of an algorithm in target, so the model is trained to generate not only the final answer, but also the intermediate steps, which is similar to chain-of-thought, but requires more training data. \citet{scratchpad} show that using the training data augmented with intermediate steps significantly improves the model performance, but even with 100k augmented training examples for the addition task, the fine-tuned 1B LM still does not perform well on out-of-distribution addition.

Our work is also related to \citet{ntm}, which extends the capabilities of Recurrent Neural Networks to two simple symbolic manipulation tasks, copy and sort, by augmenting the model with external memory resources. Instead of using hundreds of thousands of training examples, we focus on large pretrained LMs with few-shot prompting or fine-tuning settings in this work.





\section{Observations}
\label{sec:observation}
\begin{figure}[!h]
\centering
\includegraphics[width=0.4\textwidth]{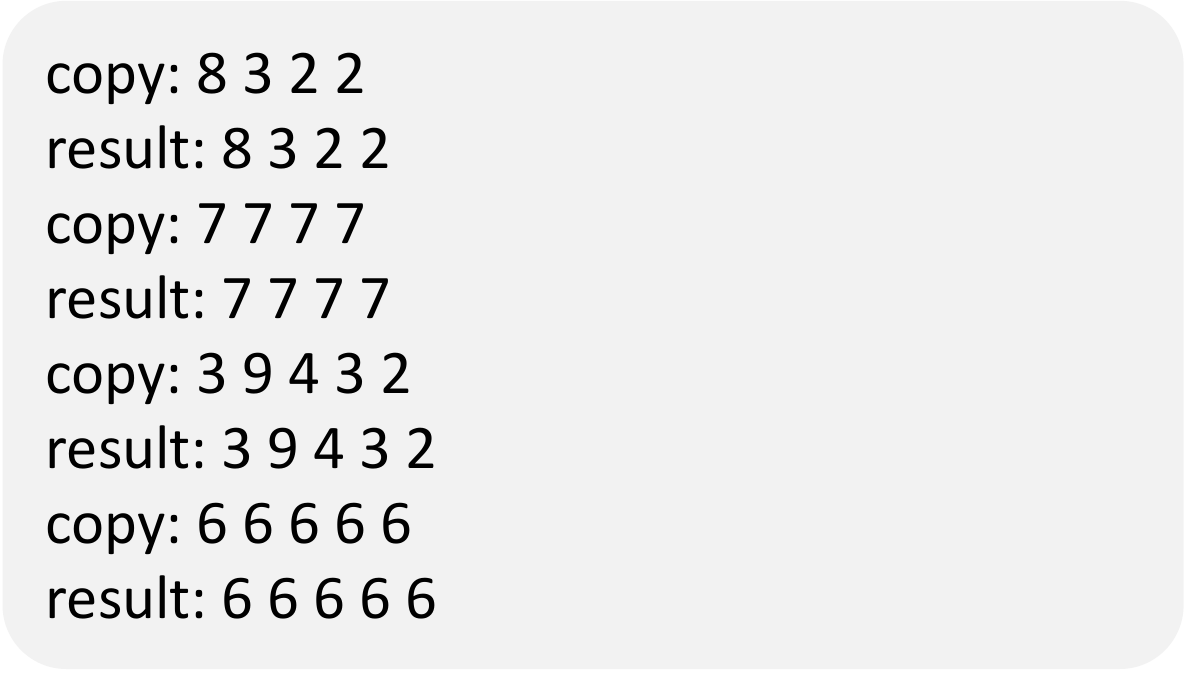}
 \caption{The prompt for GPT3 on the copy task.}
\label{fig:gpt3_copy_prompt}
\end{figure}

We first analyze the difficulty of generalizing the copy operation, one of the most fundamental, simplest symbolic manipulation operations.
We start by copying random numbers. For GPT3, we augment each testing example with the few-shot prompt as shown in Figure \ref{fig:gpt3_copy_prompt}. 

We also fine-tune a T5 model for copying. The training data follows the same format as above and consists of random numbers of up to 5 digits. 
We first evaluate the prompted GPT3 and fine-tuned T5 on copying random numbers of up to 80 digits ($\alpha=0.1$ in Figure~\ref{fig:intro-repeat}). GPT3 achieves nearly 100\% accuracy on all the testing examples of up to 80 digits when $\alpha$ is 0.1. 
The finetuned T5 does not generalize well beyond 7 digits, and it achieves nearly 100\% accuracy on all the testing examples of 7 digits,  except for a few error cases as follows:\\
\texttt{
``input'': copy: $\cdots$ 9 8 9 8 9 4 $\cdots$ \\
``pred'': $\cdots$ 9 8 9 4 $\cdots$  \\
``input'': copy: $\cdots$ 6 0 6 0 6 5 $\cdots$ \\
``pred'': $\cdots$ 6 0 6 5 $\cdots$ \\
}
A common feature of these error cases is that they all contain consecutive repeating numbers and the model tends to mistakenly skip part of them or over-replicate them. 

\begin{figure}[!t]
\centering
\includegraphics[width=0.47\textwidth]{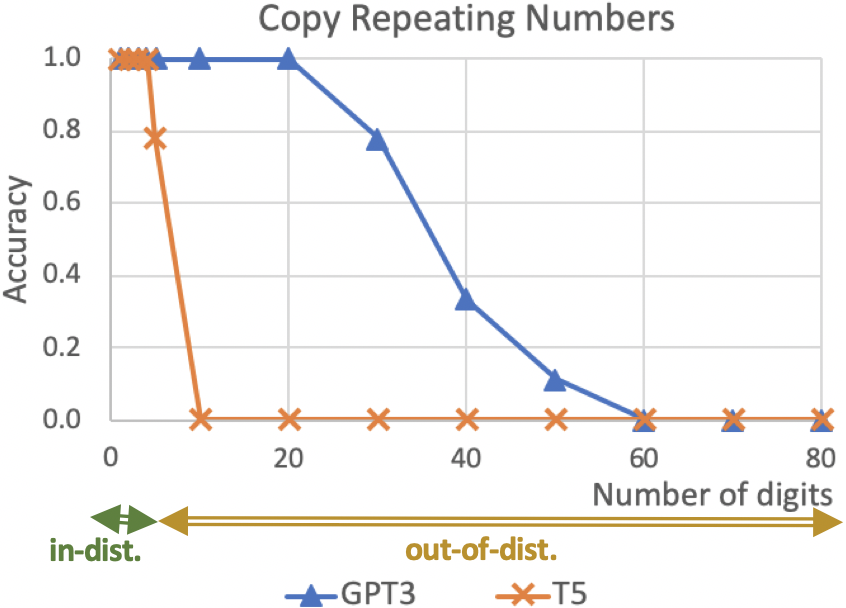}
\caption{An illustration of GPT3 and T5 performance on copying repeating numbers. }
\label{fig:copy-repetitive-baselines}
\end{figure}

Therefore we further do a copying task where the testing examples consist only of repeated digits, such as \texttt{copy: 2 2 2 2 2}.
The results are shown in Figure~\ref{fig:copy-repetitive-baselines}. Although T5 performs well at copying random numbers of up to 7 digits, its accuracy at copying 5-digit repeating numbers drops below 80\%. Similarly, the prompted GPT3 can not generalize to copying more than 30 repeating digits. Both GPT3 and T5 are Transformer-based LMs, which use the self-attention mechanism. When copying numbers, the models are required to use self-attention to locate the next digit to copy. When copying random numbers without repeated digits, it usually would be sufficient to locate the next digit by comparing the previous few digits. However, when copying repeated digits, this mechanism no longer works since all the previous digits are the same. Instead, the model needs to locate the next digit either by counting the repetitive digits or by remembering the previous position. Therefore, the results suggest that the Transformer-based LMs, such as GPT3 and T5, might have limited capability of locating in symbolic manipulation.

\section{Mitigation Methods}
\label{sec:method}
In order to mitigate the limitations, we examine a few potential solutions.

\subsection{Positional Markers}
\label{subsec:method-implicit}
We first explore possible methods to mitigate the problem of repeating numbers. We introduce two types of positional markers: implicit positional markers and explicit ones.

\begin{figure}[!t]
\centering
\includegraphics[width=0.47\textwidth]{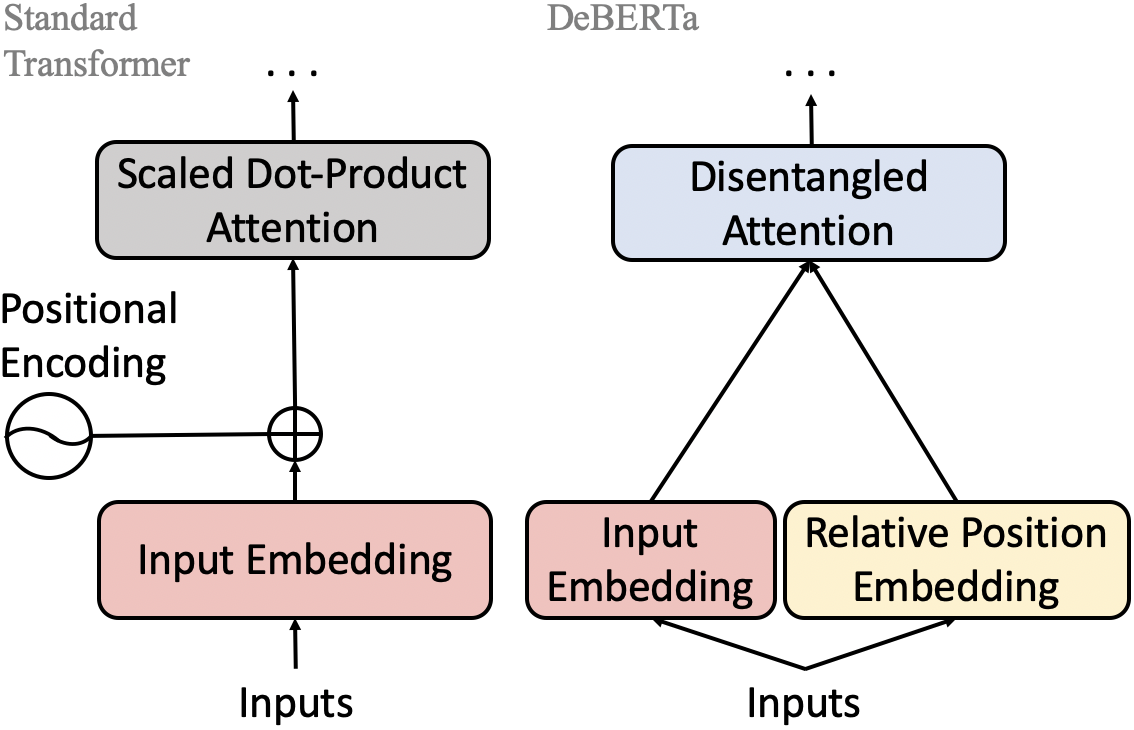}
\caption{An illustration of standard Transformer attention (left) and DeBERTa disentangled attention (right). }
\label{fig:deberta}
\end{figure}

As stated above, LMs tend to make mistakes when the input contains repeating numbers. When humans deal with repeating numbers in basic arithmetic, we usually use indices to distinguish those digits in the input. Positional encoding in large LMs is closely related to this human practice. Most Transformer-based LMs encode the positional information into positional vectors and add each of them to the corresponding word vector. Although large LMs have already incorporated positional encoding in the model architecture (Figure~\ref{fig:deberta}), results in Figure~\ref{fig:intro-repeat} indicate that the positional encoding commonly used in large LMs may not be sufficient to locate each repeating digit effectively. 

Instead of representing each token by the sum of its contextual token embedding and the position embedding, DeBERTa~\cite{deberta} represents each token with a token embedding and a position embedding, respectively, and the attention weights are computed using disentangled matrices based on both embeddings, respectively (Figure~\ref{fig:deberta}). In other words, the self-attention in DeBERTa is disentangled. With the disentangled relative position embeddings, the attention scores between tokens depend not only on the content but also on the relative position between the tokens, so the disentangled relative position embeddings act as implicit position markers within DeBERTa, which might make it easier for the model to learn the latent position relationship in the training data of the symbolic manipulation tasks. 

Although DeBERTa uses disentangled attention mechanism, it was not originally introduced to enhance the locating capability of LMs, so no pretraining task was specifically proposed for training the position embeddings in DeBERTa. This may potentially lead to limited generalization ability of DeBERTa on the induction tasks requiring accurate locating.

Rather than relying on implicit positional markers, another, more straightforward approach is to add explicit positional markers in the input for the model. For example, the input string \texttt{2 2 2} is augmented with positional markers \texttt{A, B, C, $\cdots$}. We explore two methods of adding explicit positional markers:\\
\textbf{Ordered marker}: The markers are inserted into the input in order. \texttt{2 2 2} $\rightarrow$ \texttt{A 2 B 2 C 2}\\
\textbf{Random marker}: The markers are inserted into the input in random order. \texttt{2 2 2} $\rightarrow$ \texttt{E 2 X 2 J 2}  

With the explicit positional markers, each repeating \texttt{2} becomes different for the model. When doing symbolic manipulation, the Transformer-based LMs can easily locate the digit by recognizing the explicit positional markers. 
Essentially, adding explicit positional markers breaks the repeating numbers into a non-repeating input sequence. This method is also related to pointer networks \cite{ptrnet}, which uses attention as a pointer to select the position indexes of the input tokens as the output.  A hybrid pointer-generator network can also be leveraged to copy number from the source text, while retaining the ability to produce new numbers through the generator \cite{see-etal-2017-get}.   Compared with implicit markers, explicit markers provide more direct and clearer location information in text format. However, similar to the implicit positional markers, whether using the explicit positional markers can generalize to arbitrary length or unseen markers is still questionable. 



\subsection{Fine-grained Computation Steps}
\label{subsec:method-cot}
We then explore possible methods to alleviate the OOD generalization problem.
One observation is that the complexity of addition with long digits is larger than that of the 1-digit addition. Thus, the model should be given more computation time on the task when the numbers are large. The fine-tuned T5 and prompted GPT3 mentioned above, however, is required to generate the answer with a fixed amount of computation, so one possible direction to mitigate this limitation is to allow the model to operate step-by-step instead of generating the answer in one forward pass. For example, in k-digit addition, the model is allowed to break it down into k simple 1-digit addition and the model is allowed to generate k intermediate addition results to get the final answer. 

Generating fine-grained computation steps can potentially alleviate the generalization problem, but may not contribute to the locating capability of the Transformer-based LMs. To mitigate the locating problem, we add positional markers to scratchpad~\cite{scratchpad} (Figure \ref{fig:gpt3_addition_prompt}):

\begin{figure}[!h]
\centering
\includegraphics[width=0.48\textwidth]{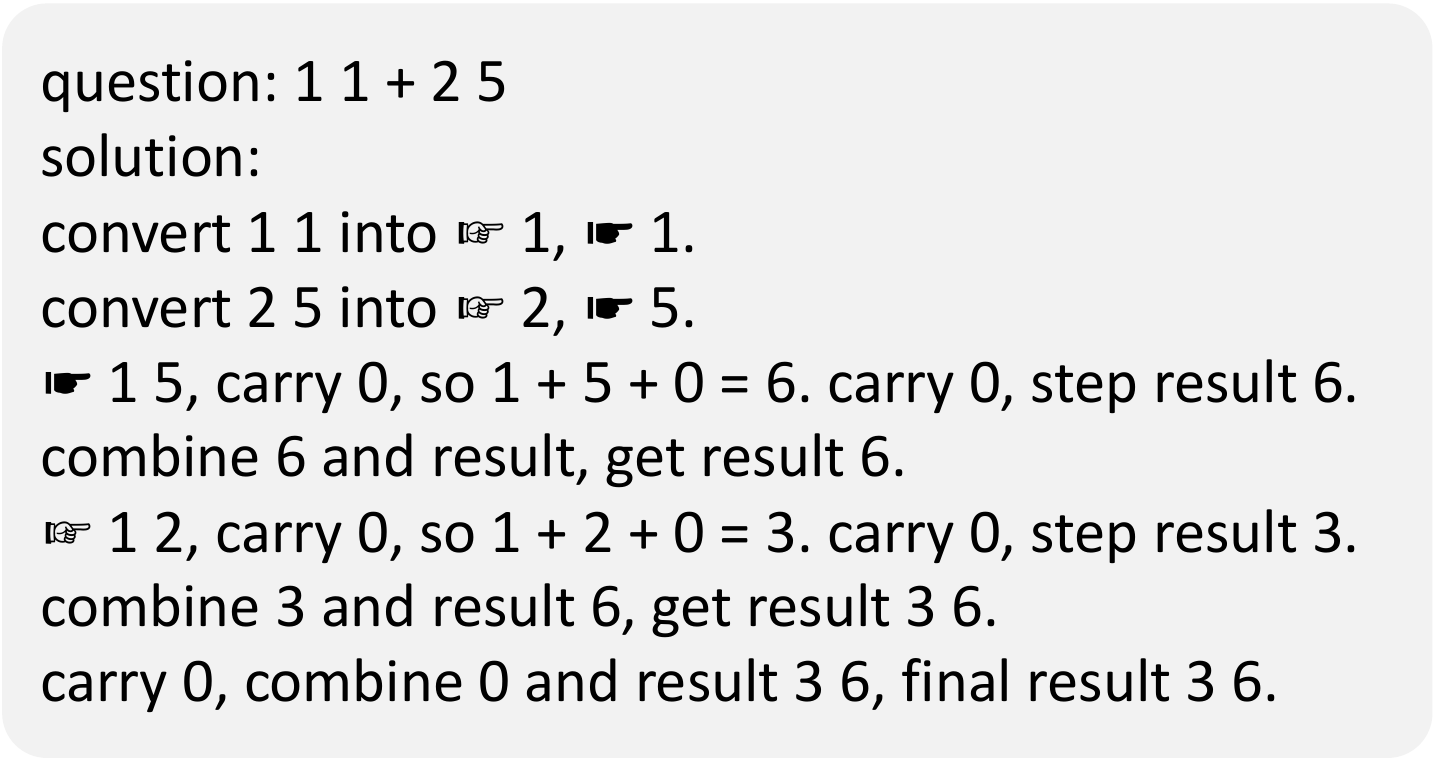}
 \caption{The prompt for GPT3 on the  addition task. We use \faHandPointRight[regular] and \faHandPointRight ~to denote optional different markers as described in Section \ref{subsec:method-implicit} if they are applied.}
\label{fig:gpt3_addition_prompt}
\end{figure}

\begin{figure}[!h]
\centering
\includegraphics[width=0.46\textwidth]{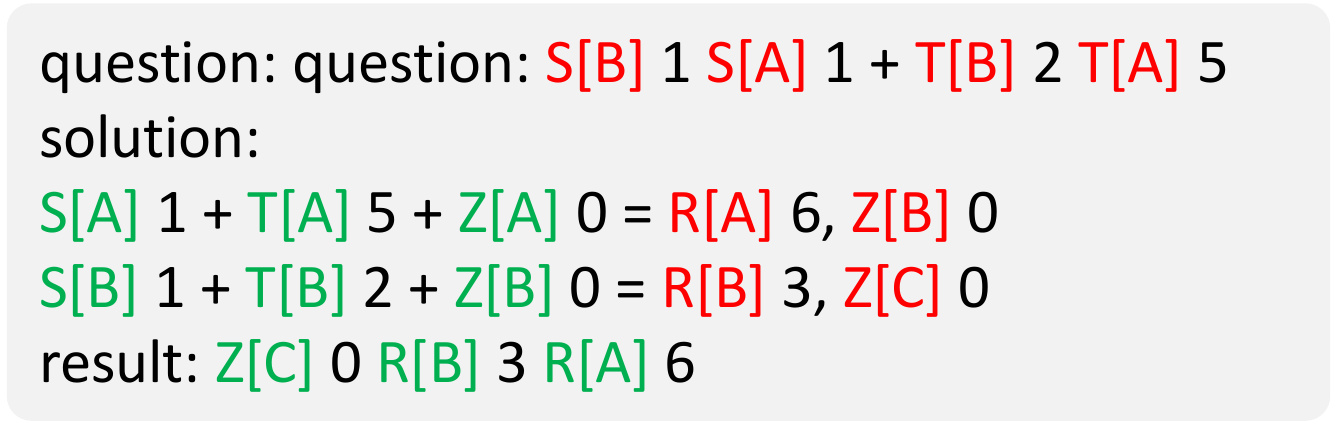}
 \caption{The demonstration of comprehensive scheme for addition problem, where position marker is marked red and reference marker is marked in green.}
\label{fig:addition_reference_marker}
\end{figure}

We also experiment a more comprehensive scheme where each number in the demonstration is associated with an explicit positional marker or reference marker.  A reference marker refers the positional marker where the following number is copied from as shown in Figure \ref{fig:addition_reference_marker}.

Through our experiments, we found that although these markers can help achieve higher accuracy for in domain data. It does not help much for OOD data. It clearly indicates the limitation of Transformers and pre-trained language models in induction. In the following discussion, we will shed some insights on how to eliminate such limitation.


\subsection{LM with callable Programs}
\label{subsec:method-callable}
If both implicit positional markers and explicit positional markers do not generalize well in a symbolic reasoning task, then an alternative is to combine LMs with callable programs to replace the basic symbolic operations when possible, since callable programs do not have the generalization problem. For example, when combined with the fine-grained computation steps in the addition task, the convert, add, or combine operations can be considered callable programs. When the LM generates the text sequence \texttt{add(1,5)}, the callable function \texttt{add} will be invoked and return the result in text: \texttt{carry C: 0, result 6}.

Following the example in Section~\ref{subsec:method-cot}, with callable functions, the prompt format is as follows:
\begin{figure}[!h]
\centering
\includegraphics[width=0.48\textwidth]{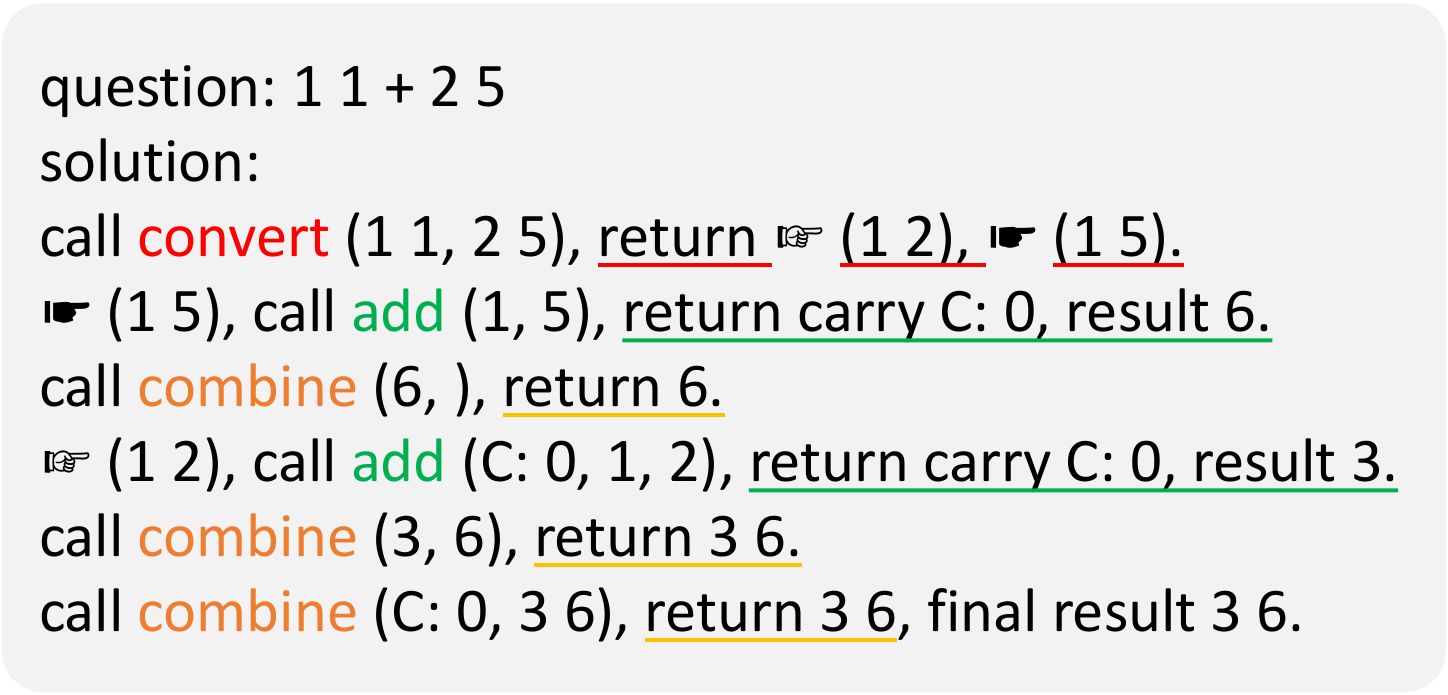}
 \caption{The prompt for GPT3 on the addition task with callable programs. \faHandPointRight[regular] and \faHandPointRight ~are positional markers. Different callable programs (convert, add and combine) are marked in different colors, and the results they returned are underlined with the corresponding color.}
\label{fig:gpt3_callable_program}
\end{figure}


Given a testing example, the prompted GPT3 first generates the solution step by step. During the process, the results of the function calls will be appended to the generated result to be used in the following steps.

Callable programs can be viewed as decomposing a complex task to smaller, simpler jobs. The remaing issue is to learn chaining these smaller jobs together to complete the task.

\begin{figure}[!h]
\centering
\includegraphics[width=0.48\textwidth]{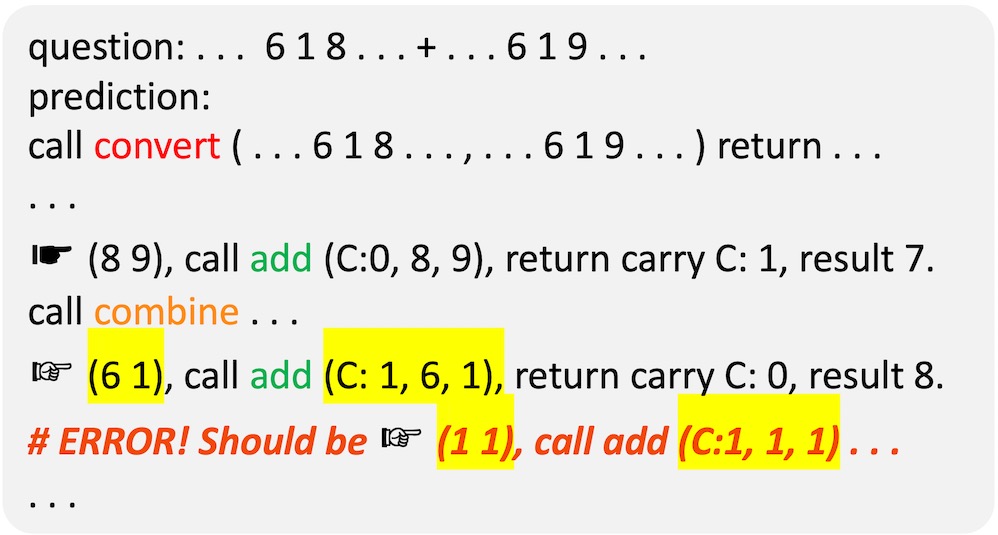}
 \caption{An error example of GPT3 with callable functions. The error is highlighted.}
\label{fig:gpt3_callable_program_error}
\end{figure}

Callable programs can guarantee the correctness of output given correct input for a given job. However, even augmented with callable programs, LMs may still suffer from the locating problem since the callable programs rely on LMs to decide which token to copy (Figure~\ref{fig:gpt3_callable_program_error}). Unfortunately, LMs cannot guarantee the correctness of this copy action. 

\subsection{LM with Tutor}
\label{subsec:method-simulator}

Scratchpad \cite{scratchpad} ignores the visual process when an elementary school tutor visually illustrates how to perform addition step by step: Pinpointing where each digit in the output sequence comes from, adding single digits together and iterating.  It turns out that these details and abstractions are important in order to simplify the learning process and help kids learn how to do addition in a few shots.

A tutor shows every single step visually and sometimes calls an already learned sub-module to complete a task. In this way, the hypothesis space between two consecutive steps can be dramatically simplified; hence the chance of learning a correct model  can be improved.  

Taking copy as an example.  Instead of providing a training example: \texttt{copy:  1 1 1 2 2 2 result: 1 1 1 2 2 2}, we need to demonstrate where the first \texttt{1}, the second \texttt{1}, the third \texttt{1} in the output sequence comes from, which exactly imitates the finest action a human could do to perform such an operation.  Suppose there is a cursor placed at the beginning of the input sequence, a ``rmov'' operation moves the cursor one token to the right. A ``cpy'' operation copies a single digit to the output sequence.  An ``end'' operation checks if the marker reaches the end of the sequence. ``T'' and ``F'' represent true and false respectively. We assume all these actions are unitary and have been learned.  Then a possible action sequence to complete the copy operation is as follows: \\
\texttt{rmov, end=F, cpy, rmov, end=F, cpy, \ldots, rmov, end=T.}\\
\begin{figure}[!t]
\centering
\includegraphics[width=0.37\textwidth]{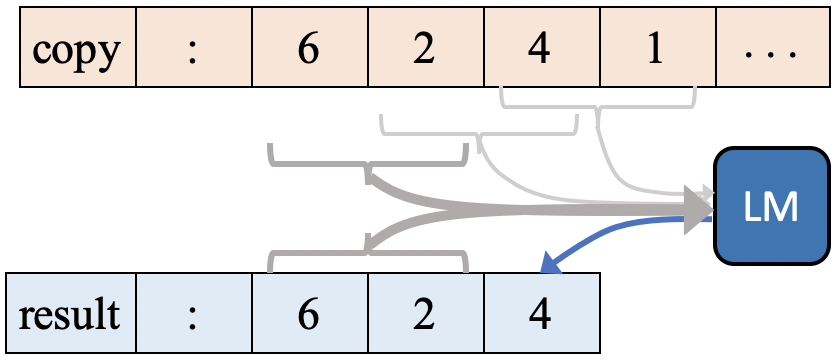}
\caption{An illustration of doing copy with pattern matching.}
\label{fig:pattern}
\end{figure}
This fine-grained action sequence accurately describes the whole copy operation. Certainly, there are other ways to perform copying. For example, instead of using a cursor, one can use a pattern match to perform the copy operation (Figure~\ref{fig:pattern}).  
We suspect that the copy operation learned from Transformer is following this pattern-matching approach, which is error-prone when the pattern has repeating symbols and when the long pattern is out-of-distribution.  Positional markers do not help either as it seems unable to handle the OOD generalization problem.

If we take the action sequence ``rmov, end=F, \ldots'' to train a Transformer for copying, the hypothesis space is  simplified, thus making it possible to find the simplest model that can simulate the whole action sequence. This is related to imitation learning \cite{alvinn,imitationross}.  Although there is no guarantee that Transformer can definitely find the correct model, the chance is much higher.  One can also relate the setting with a multiple tape Turing machine where state transition is conducted among the positions of tape heads and read/write operations.   The Transformer is trained to learn such state transition, thus completing the programming of a Turing machine. 

\begin{figure}[!t]
\centering
\includegraphics[width=0.47\textwidth]{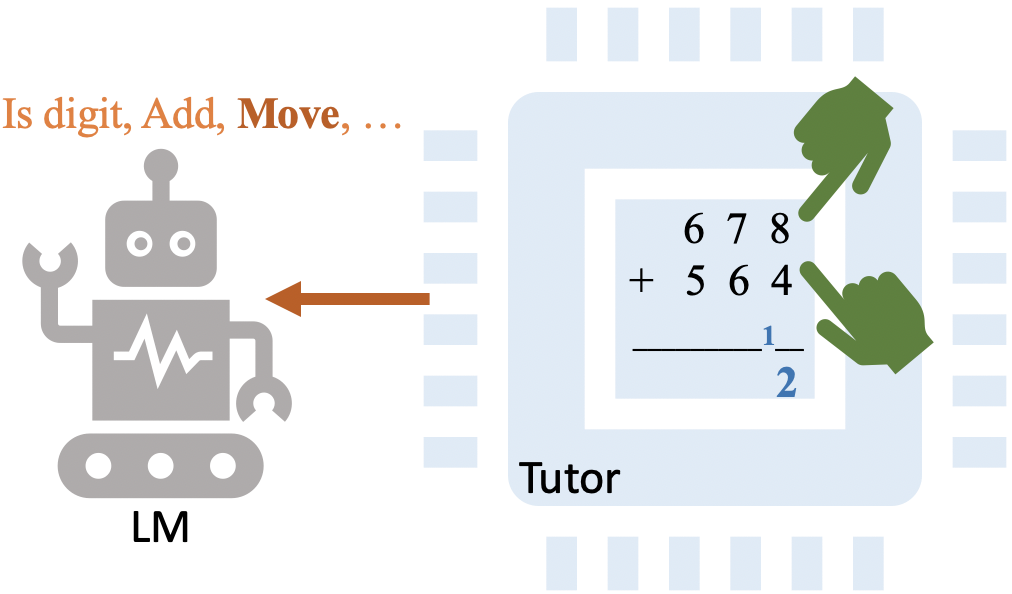}
\caption{An illustration of the LM with Tutor method. With the tutor (right), the LM (left), or just a transformer  generates an action sequence that simulates how humans do arithmetic addition.}
\label{fig:simulator}
\end{figure}

As for the addition operation, a similar action sequence can be obtained to simulate how humans tutor kids do addition at an early age (Figure \ref{fig:simulator}). Let ``lmov'' denote moving the cursor one token to the left.  The ``add'' operation adds three single digits together, one from each of the two operands and the third one from the carry digit, appends the result to the output, and updates the carry digit. Assume ``add'' is a callable program as kids have learned how to do single digits addition.  Suppose the cursor starts from the end of the operands. The entire action sequence looks like the following. \\
\texttt{lmov, end=F, add, lmov, end=F, add, \ldots, lmov, end=T.}

The main difference between the tutor and the Scratchpad method \cite{scratchpad} is the abstract callable function and detailed action sequence.   The action sequence includes all the state transitions needed to complete the task.  It perfectly overcomes the OOD issue and does not require many training examples in order to achieve 100\% accuracy. 

While there is a great effort to enlarge Transformer-based LMs such as PALM \cite{palm} and Minerva \cite{minerva}, to improve the performance in symbolic and logical reasoning, our result reveals that it might be necessary to demonstrate the action sequence with reasonable abstraction to the Transformer to leverage its full strength. 

The action sequence ``rmov, end=F, ...'' can also be viewed as an algorithm execution instance.  It is well known that if one step is missing in an algorithm, it will most likely not produce the expected output.  In order to fill the gap in case one step is missed during demonstration, we have to rely on training examples of that step and learn it,  which will likely incur errors.  As the number of missing steps increases, errors will accumulate and eventually make  learning more difficult and hard to generalize.  To make learning in symbolic reasoning easier, it will be important  to have a detailed action sequence as we have demonstrated in Figure~\ref{fig:simulator}.

\section{Experiments}

In this section, we conduct experiments on three different problems including copying, addition, and another basic symbolic manipulation operation, reverse. We illustrate the limitation of LMs in symbolic and arithmetic induction and the improvement that could be achieved by the introduced mitigation methods.


\begin{figure*}[!t]
\centering
\includegraphics[width=0.97\textwidth]{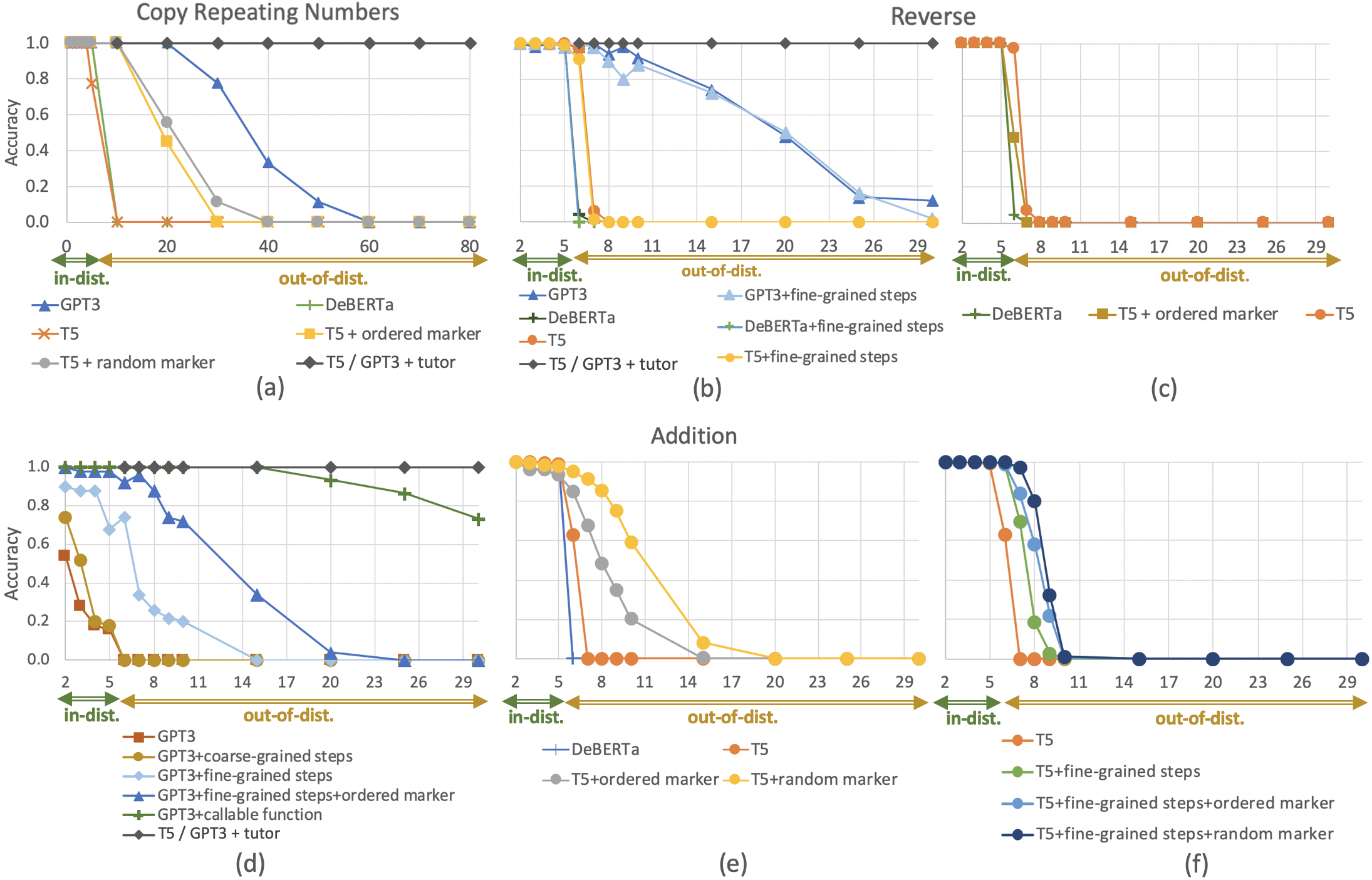}
\caption{Experimental results. (a): results of copying repeating numbers. (b)(c): results of reversing the list. (d)(e)(f): results on arithmetic addition. The x-axis is the number of digits or number of items.}
\label{fig:results}
\end{figure*}

\subsection{Copy Operation}
\label{subsec:exp-copy}
Copying is the most basic operation used in symbolic manipulation and arithmetic reasoning. We experiment with the following methods:\\
\textbf{GPT3}: We use the prompt as shown in Section~\ref{sec:observation}.\\
\textbf{DeBERTa / T5}: The training example is as follows: \texttt{copy: 1 2 3 4  result: 1 2 3 4}\\
\textbf{T5 + ordered marker}: The training data is augmented with explicit positional markers. \texttt{copy: A 1 B 2 C 3 result: A 1 B 2 C 3} \\
\textbf{T5 + random marker}: Same as above, but the augmented positional markers are in random order. \texttt{copy: E 1 A 2 F 3 result: E 1 A 2 F 3} \\
\textbf{T5 / GPT3 + tutor}: The training and testing examples are as described in Section~\ref{subsec:method-simulator}.

We experiment with the T5-base (220M) model, DeBERTa-base (140M) model, and GPT3 text-davinci-002. The models are initiated with the pretrained parameters and further fine-tuned on the training data. For GPT3 or T5 with tutor, the training data consists of 15 examples of up to 5 digits. For all the other T5 models and DeBERTa, the training data consists of 2,000 random numbers of up to 5 digits. We evaluate all the models on copying repeating numbers of up to 80 digits. The results are illustrated in Figure~\ref{fig:results}(a). 

As shown in Figure~\ref{fig:results}(a), GPT3 achieves 100\% accuracy on the in-distribution testing data (1-5 digits) but the fine-tuned T5 achieves 78\% accuracy on the 5-digit repeating numbers although they are in-distribution. Augmented with random or ordered positional markers, the T5 models achieve 100\% in-distribution accuracy, and so does using implicit positional markers (DeBERTa). This suggests that both implicit positional markers and explicit positional markers may help with the locating capability of LMs. However, using explicit positional markers, either ordered or random, the model exhibits significantly better generalization to OOD testing data whereas DeBERTa fails on OOD data. GPT3 exhibits better OOD generalization than T5 with positional markers but it does not generalize well beyond 30 digits. Both \textit{T5 + tutor} and \textit{GPT3 + tutor} keeps 100\% accuracy when the number of digits increases.


\begin{figure}[!th]
\centering
\includegraphics[width=0.45\textwidth]{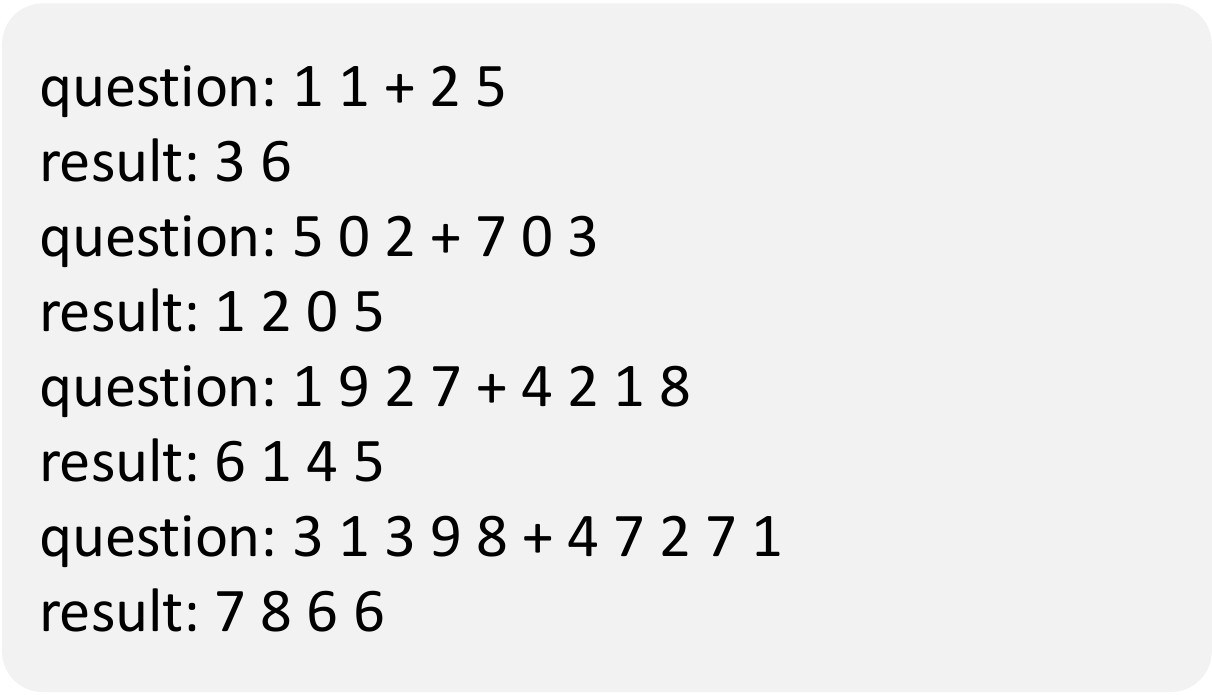}
 \caption{The prompt for GPT3 on the addition task without intermediate steps.}
\label{fig:gpt3_addition_wo_step}
\end{figure}

\subsection{Addition}
\label{subsec:exp-addition}
For arithmetic addition, we experiment with the following methods:\\
\textbf{GPT3}: The exemplars are shown in Figure \ref{fig:gpt3_addition_wo_step}.\\
\textbf{GPT3 + coarse-grained steps}: The exemplar is similar to that in Figure~\ref{fig:gpt3_addition_prompt}, but the instructions for the result combination and the computation of the carry digit and step result are omitted.\\
\textbf{GPT3 + fine-grained steps (+ ordered marker)}: The exemplar we use is as shown in Figure~\ref{fig:gpt3_addition_prompt}. \\
\textbf{GPT3 + callable programs}: The exemplar is shown in Figure~\ref{fig:gpt3_callable_program}.\\
\textbf{DeBERTa / T5}: The training data follows the format of the exemplar for GPT3. \\
\textbf{DeBERTa / T5 + fine-grained steps}: The training data used in this setting follow the format as the exemplar in \textit{GPT3 + fine-grained steps}.\\
\textbf{T5 + ordered / random marker}: The training example is augmented with ordered or random markers. For example, \texttt{question: G 1 C 1 + G 2 C 5 result: G 3 C 6}.\\
\textbf{T5 + fine-grained steps + ordered / random marker}: The training data in this setting follow a similar format as the exemplar in \textit{GPT3 + fine-grained steps + ordered marker}, but the positional markers can be in random order.\\
\textbf{T5 / GPT3 + tutor}: The training and testing examples are as described in Section~\ref{subsec:method-simulator}.

The model settings are the same as in the above copy experiments. For LMs with tutor, the training data or prompt consists of 15 examples of up to 5 digits. In other settings, the training data consists of 1,000 examples of 1-5 digit addition and for GPT3, the prompt includes 4 examples. We evaluate all the models on the addition of up to 30 digits. The results are shown in Figure~\ref{fig:results}(d)(e)(f).


As shown in Figure~\ref{fig:results}(d), both coarse-grained and fine-grained computation steps contribute to the in-distribution performance of GPT3, and using finer-grained steps achieves larger performance gains on both in-distribution data and OOD data. The performance is further boosted with explicit positional markers. Experiments on T5 (Figure \ref{fig:results}(e)(f)) also show the effectiveness of using explicit positional markers, with or without fine-grained computation steps, indicating that the explicit positional markers might make it easier for LMs to learn the induction in the arithmetic reasoning tasks. 
Similar to the results on the copying task, both DeBERTa and \textit{DeBERTa + fine-grained steps} achieve near 100\% in-distribution accuracy but 0\% OOD accuracy, suggesting that the relative position embedding of DeBERTa might have limited OOD generalization ability.
On T5, incorporating fine-grained computation steps does not improve the OOD performance as significantly as on GPT3 (Figure~\ref{fig:results}(f)). The reason might be that fine-tuning T5 tends to overfit more easily than prompting GPT3.
Unsurprisingly, \textit{GPT3 + callable programs} achieves much better OOD generalization. However, its OOD performance still degrades as the number of digits increases. Same as in the copy experiments, \textit{LMs + tutor} keeps 100\% accuracy on all the experimented numbers of digits.



\subsection{Reverse List}
\begin{figure}[!h]
\centering
\includegraphics[width=0.45\textwidth]{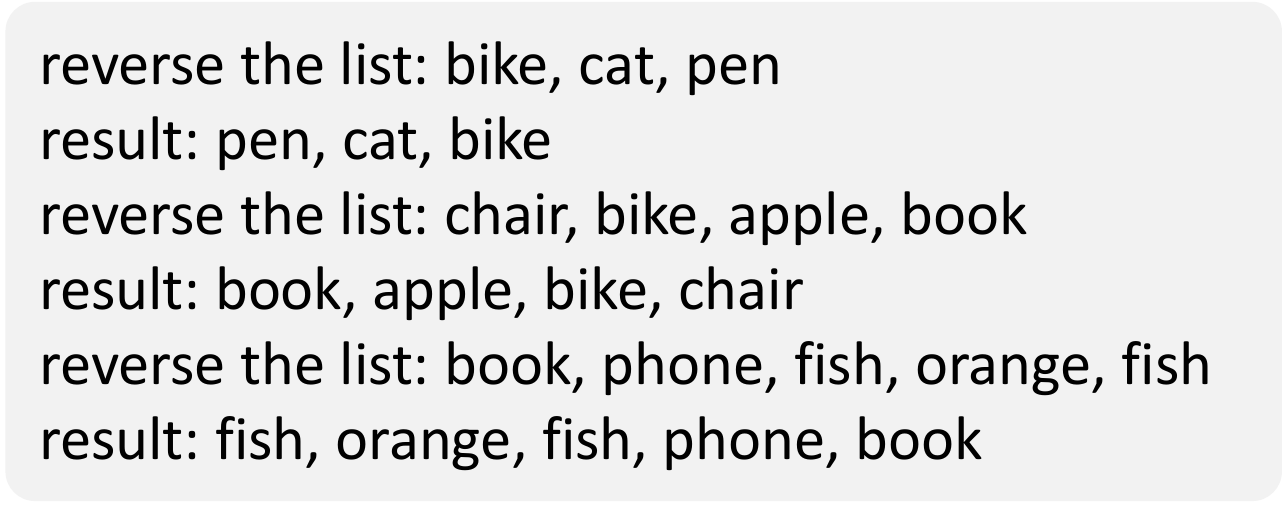}
 \caption{The prompt for GPT3 on the reverse task without intermediate steps.}
\label{fig:gpt3_reverse_wo_step}
\end{figure}

Besides copying and addition, we also experiment with reversing. Reversing is similar to copying. Both require replicating the items in the input, but reversing might be more challenging than copying in the terms of locating. In copying, the distance between each source digit and the replicated digit is the same for each digit in the number. However, when reversing, the distance between the source item and the replicated item keeps increasing during the generation. 
For this problem, we experiment with the following methods: \\
\textbf{GPT3}: The prompt without any intermediate steps is used, as shown in Figure \ref{fig:gpt3_reverse_wo_step}.

\begin{figure}[!h]
\centering
\includegraphics[width=0.45\textwidth]{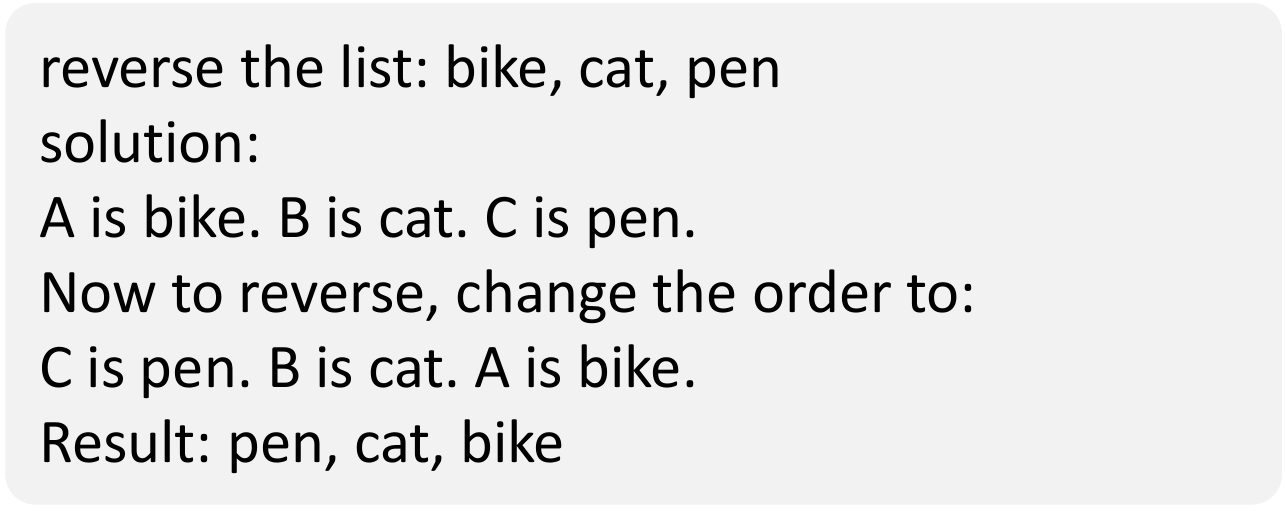}
 \caption{The prompt for GPT3 on the reverse task with fine-grained steps.}
\label{fig:gpt3_reverse_with_fine_grained_step}
\end{figure}

\noindent
\textbf{DeBERTa / T5}: \texttt{reverse the list: bike, apple, book result: bike, cat, pen} \\
\textbf{GPT3 / DeBERTa / T5 + fine-grained steps}: The training example for T5 and the exemplar for GPT3 are shown in Figure \ref{fig:gpt3_reverse_with_fine_grained_step}.

\noindent
\textbf{T5 + ordered marker}: The list items are augmented with the ordered positional markers in the input. \texttt{reverse the list: A bike, B cat, C pen result: pen, cat, bike}. \\
\textbf{T5 / GPT3 + tutor}: The training and testing examples are very similar to that for the copy task. The only difference is the direction of the move operation. ``rmov'' in the copy task is replaced by ``lmov'' here.

The model settings are the same as in the above experiments and the training data consists of examples of 1-5 items, which are randomly sampled from a predefined list of single-token nouns. For LMs with tutor, the training data or prompt consists of 15 examples of up to 5 items. In other settings, the training data consists of 1,000 examples for T5, and for GPT3, each prompt includes 4 examples. We evaluate all the models on reversing the list of up to 30 items. The results are shown in Figure~\ref{fig:results}(b)(c).


Although GPT3 can generalize to 80 digits on copying random numbers (Figure~\ref{fig:intro-repeat}), it does not generalize well beyond 20 items on reversing, which suggests that reversing might require stronger locating capability than copying. This problem also occurs on DeBERTa and T5. When tested on the OOD data, the models tends to generate only a sublist of the input. Using fine-grained steps (Figure~\ref{fig:results}(b)) or positional markers, whether implicit or explicit (Figure~\ref{fig:results}(c)), does not significantly improve the generalization of the experimented models. The reason might be the increasing distance between the source item and the replicated item as stated above. Again, \textit{LMs + tutor} maintains 100\% accuracy throughout the experiments.

\subsection{Discussion}
\label{subsec:discussion}
From the experimental results, we observe that fine-grained computation steps may improve the LM's induction ability on the arithmetic reasoning tasks and the granularity of the steps has an impact on the performance improvement. Finer-grained computation steps may contribute to larger performance improvement. 

Positional markers, whether implicit or explicit, improves LMs' in-distribution performance on all the symbolic manipulation tasks in our experiments. However, We find that augmented with the relative position embeddings, DeBERTa tends to face more severe over-fitting than T5 during fine-tuning. In the reversing experiment, using the T5 model without pretrained parameters, the fine-tuned model can not achieve a good in-distribution performance after 200k optimization steps. However, the DeBERTa model without pretrained parameters achieves 100\% in-distribution accuracy within only 2k optimization steps while the OOD accuracy drops, indicating that it has overfitted within 2k optimization steps. In other words, the relative position embeddings in DeBERTa significantly improve the model's capacity of positions, which improves in-distribution performance on simple symbolic manipulation tasks, but may not generalize well on OOD data. Compared with the implicit positional markers (relative position embeddings in DeBERTa), explicit positional markers might have better OOD generalization ability. However, incorporating symbolic manipulation tasks in the LM pretraining stage might alleviate this problem, so incorporating implicit positional markers can still be a possible direction of improving the LM's performance on reasoning tasks requiring locating ability.

Using LM with callable programs exhibits strong OOD performance on addition, suggesting that the LMs' ability to perform simple symbolic operations, such as copying, splitting, and combining, can be critical for improving their performance on reasoning tasks. How to further improve the LMs' performance on more complex reasoning tasks in this direction is left for future work.



\section{Conclusion}
In this work, we explore the limitations of pretrained LMs on arithmetic reasoning and symbolic manipulation. We experiment with three simple symbolic manipulation tasks and show that improving the locating and induction capability of LMs can be important for further improving their performance on induction tasks.  Our method that combines abstraction and finest-grained step-by-step tutoring demonstrates its potential to generalize correctly, shedding light on possible directions orthogonal to scaling up LMs for future work in this area.


\bibliography{anthology,custom}
\bibliographystyle{acl_natbib}

\appendix

\section{Appendix}
\label{sec:appendix}
\subsection{Experiment configuration}
For fine-tuning the T5-base and DeBERTa model, we use the learning rate 5e-5, batch size 16, training epochs 200. The maximum generation length is set to 512. The checkpoints are evaluated every 1000 optimization steps. The random seed is fixed to 42. We use the implementation for HuggingFace~\cite{wolf2020transformers}.
For GPT3, we set temperature=0, top\_p=1, frequency\_penalty=0, and presence\_penalty=0.
All the experiments are conducted on  NVIDIA RTX A6000 GPUs.

\subsection{Reference marker}
As shown in Figure \ref{fig:addition_reference_marker}, we apply two different markers in the demonstration. The positional marker is used to define the value stored in the marker, while reference marker is used to explicitly copy the value from the positional marker with the same name. Each number in this demonstration is uniquely marked with positional or reference marker. For the positional marker, the model needs to generate both the marker and its value. For the reference marker, the model only needs to generate the marker and the value will be explicitly copied from its corresponding positional marker.

Similar to previous experiments on the addition problem, we train the model on 1-5 digits and test its performance on both in-domain (1-5 digits) and out-of-domain (6-10 digits) settings. The experimental results show that the model is able to achieve 100\% accuracy on in-domain data, but get 0\% accuracy on out-of-domain data. We also tried to extend the in-domain to 10 digits and get the same results that the model can solve in-domain problems, but fail to generalize to out-of-domain. 

\begin{figure}[!t]
\centering
\includegraphics[width=0.45\textwidth]{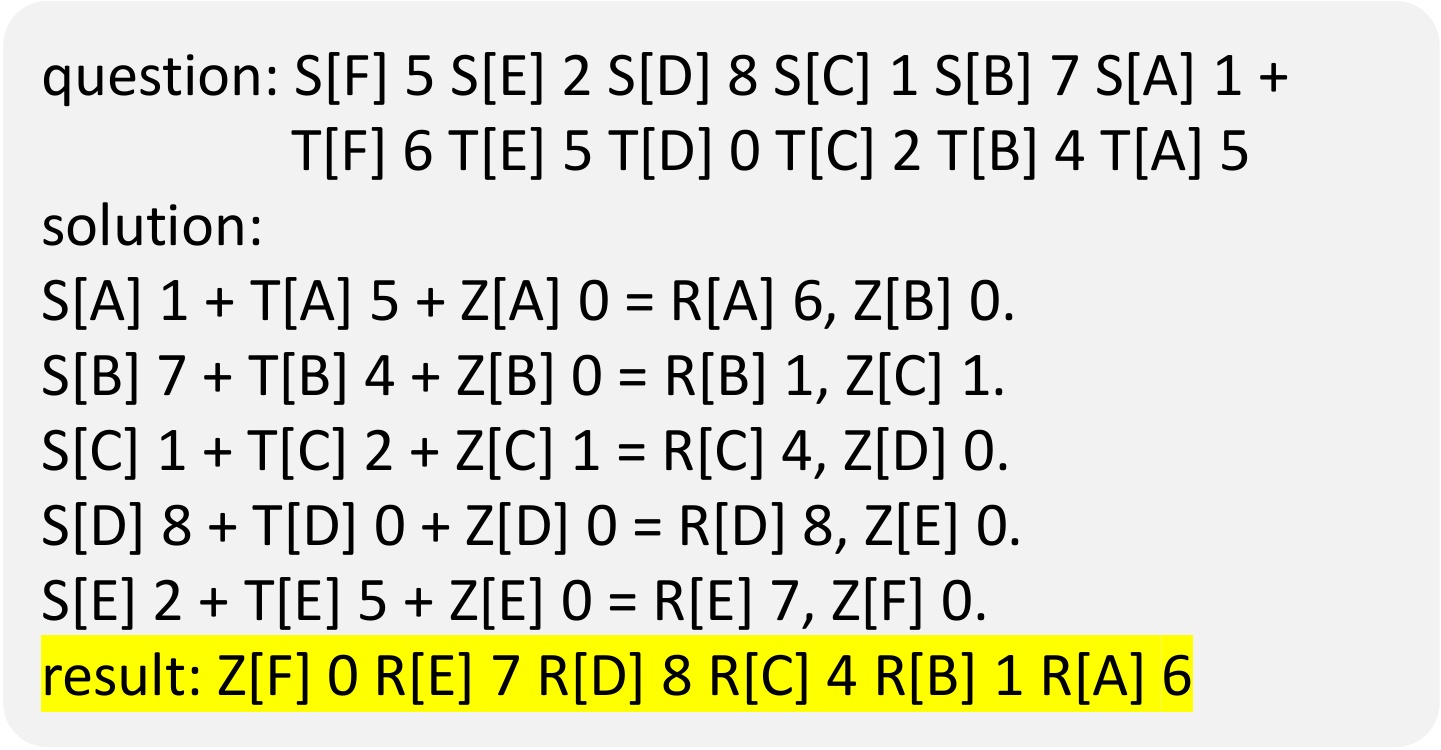}
 \caption{Error case for T5 model with positional and reference marker on addition problem.}
\label{fig:reference_marker_error_case}
\end{figure}

We show one error case of this model in Figure \ref{fig:reference_marker_error_case}, where the error step is highlighted in yellow. On this 6-digit addition problem, the model skipped the last digit and directly jump to the result, which causes the error. The problem is the model doesn't learn to how to generalize from 1-5 digits to 6 digits. Instead, it is overfitting to the training data, which makes it directly output the results after adding 5 digits. How to reduce the hypothesis space and force the model to learn to generalize to out-of-domain data would be one future research direction to solve this problem.

\end{document}